%% file: main.tex
\documentclass{article}

\usepackage{makecell}
\usepackage[final]{lg}
\usepackage[utf8]{inputenc}
\usepackage[T1]{fontenc}
\usepackage[pdfencoding=auto, psdextra, unicode]{hyperref}
\usepackage{url}
\usepackage{booktabs}
\usepackage{rotating}
\usepackage{amsfonts}
\usepackage{amsmath}
\usepackage{nicefrac}
\usepackage[dvipsnames]{xcolor}
\usepackage{adjustbox}
\usepackage{multirow}
\usepackage{amssymb}
\usepackage{longtable}
\usepackage{comment}
\usepackage[english]{babel}
\usepackage[autostyle]{csquotes}
\usepackage{hhline}
\usepackage{tabu}
\usepackage{tablefootnote}
\usepackage{graphicx}
\usepackage{wrapfig}
\usepackage{array}
\usepackage{caption}
\usepackage{tcolorbox}
\usepackage{relsize}
\usepackage{colortbl}
\tcbuselibrary{breakable}
\usepackage{floatpag}
\floatpagestyle{plain}
\usepackage{float}
\usepackage{algorithm}
\usepackage{algpseudocode}
\usepackage{threeparttable}
\usepackage{enumitem}
\setlist[enumerate,itemize]{leftmargin=1.5em}
\usepackage{placeins}

\usepackage{bm}
\usepackage{tikz}
\usetikzlibrary{positioning,arrows.meta,calc,fit,backgrounds}
\usepackage{nicematrix}

\definecolor{LightYellow}{rgb}{1.0,0.98,0.85}
\definecolor{LightPurple}{rgb}{0.957,0.918,0.987}
\definecolor{LightOrange}{rgb}{0.992,0.933,0.867}
\definecolor{Band}{gray}{0.90}
\newcommand{\cmark}{\textcolor{ForestGreen}{\checkmark}}
\newcommand{\xmark}{\textcolor{BrickRed}{\ensuremath{\times}}}

\newtcolorbox{HeroCard}{
  breakable,
  colback=black!3,
  colframe=black!3,
  boxrule=0pt,
  arc=3mm,
  left=12pt,right=12pt,top=10pt,bottom=10pt
}

\definecolor{MainPurple}{HTML}{A451E4}

\makeatletter
\edef\OrigRM{\rmdefault}
\edef\OrigSF{\sfdefault}
\edef\OrigTT{\ttdefault}
\renewcommand{\@maketitle}{%
  \begin{center}
  \begin{HeroCard}
  {%
    \fontfamily{put}\selectfont

    {\LARGE\bfseries \@title\par}
    \vspace{0.5em}

    {\normalsize \@author\par}
    \vspace{0.7em}

    {\normalsize
    \setlength{\baselineskip}{1.25\baselineskip}
    \noindent \input{sections/00_abstract}\par}

  }%
  \end{HeroCard}
  \end{center}

  \@thanks
  \global\let\@thanks\@empty
  \setcounter{footnote}{0}
  \vspace{1.25em}
}%
\makeatother

\usepackage{fourier}
\renewcommand{\rmdefault}{\OrigRM}
\renewcommand{\sfdefault}{\OrigSF}
\renewcommand{\ttdefault}{\OrigTT}

\usepackage{xspace}
\newcommand{\model}{EXAONE Finance\xspace}
\newcommand{\comp}{LG~AI~Research}

\title{
\includegraphics[height=5.5mm]{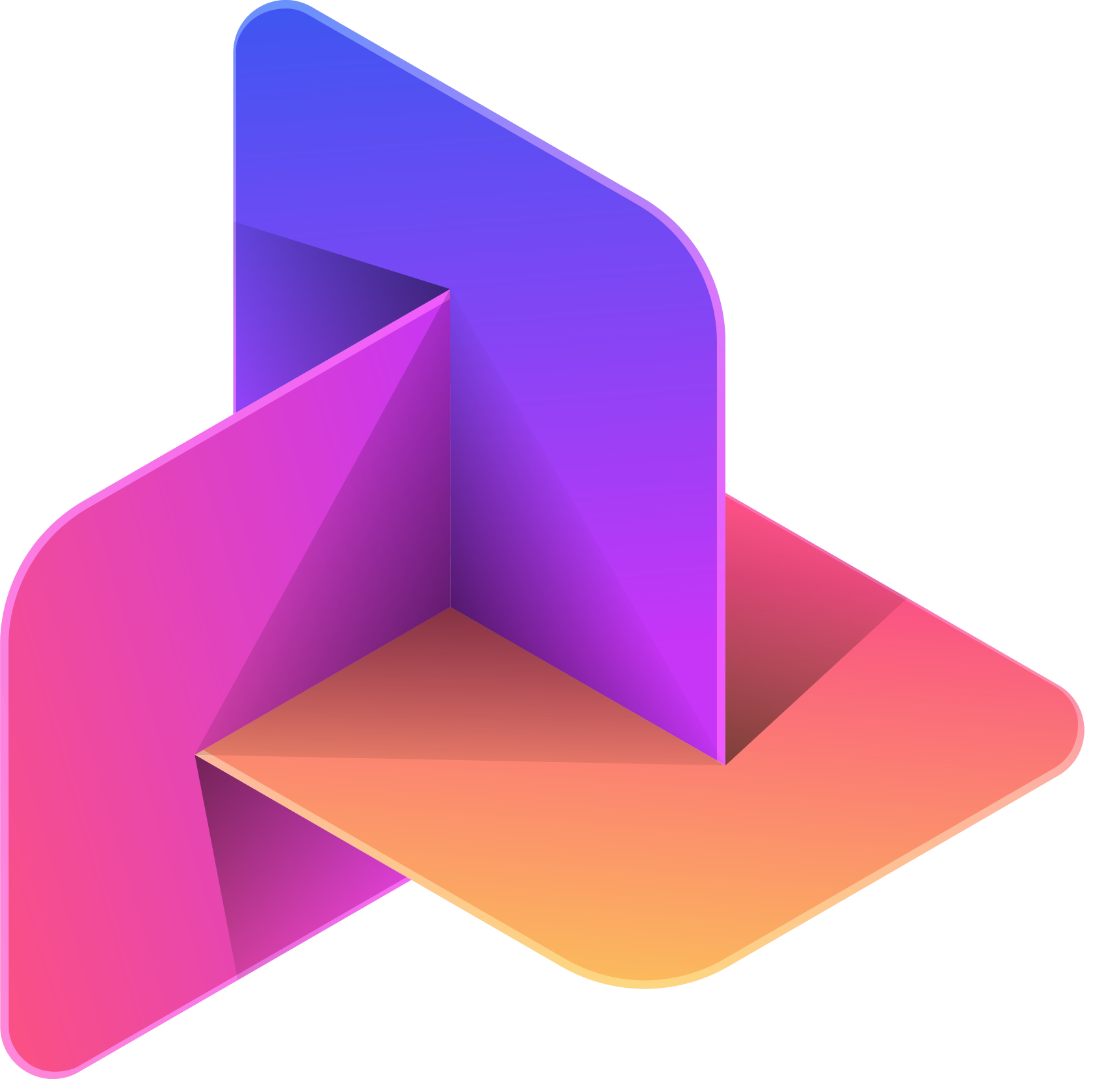}
EXAONE Finance 1.0:\\ 
\large{An Attention-free Time Series Foundation Model for Financial Time Series}
}

\author{%
  \comp$^{*}$ \\[2pt]
  {\normalsize Seunghan Lee, Jaehoon Lee, Jun Seo, Tae Yoon Lim, Dongwan Kang, Hwanil Choi, Minjae Kim, \\
  Sungdong Yoo, Junhyeok Kang, Sangjun Han, Soonyoung Lee, Wonbin Ahn}
}

\begin{document}

\maketitle

\input{sections/01_introduction}
\input{sections/02_related_work}
\input{sections/03_model}

\input{sections/04_experiments}

\input{sections/05_conclusion}

\bibliographystyle{plain}
\bibliography{refs}

\end{document}

%% file: sections/00_abstract.tex
This technical report presents \textbf{EXAONE Forecast for Finance} (EXAONE Finance), a financial time series foundation model (TSFM) 
tailored to
financial forecasting. 
While recent TSFMs achieve strong zero-shot performance through large-scale pretraining, 
they are primarily developed for \textit{general-domain} time series 
and largely rely on self-attention backbones whose computational cost grows quadratically with sequence length and variate count.
Moreover, they assume fully observed inputs and are pretrained on corpora that 
fail to adequately capture the unique dynamics of financial markets.
These limitations hinder their applicability to finance, where long, many-channel, intermittently observed panels are common.
To address these challenges, \model adopts an \textit{attention-free} architecture, replacing self-attention with two simple yet effective linear-time operators: (1) a \textit{causal 1D convolution} for temporal mixing and (2) a \textit{group-aware pooling multi-layer perceptron (MLP)} for variate mixing.
Furthermore, a \textit{masked-context augmentation} exposes the model to contiguous missing spans during training, improving robustness to the missingness pervasive in financial markets. 
\model is pretrained on a \emph{synthetic financial} corpus whose generative process is designed to reproduce the properties of financial series such as heavy tails, volatility clustering, jumps, regime shifts, and cross-asset dependence, combined with a domain-agnostic synthetic source. On FinVerse, a financial forecasting benchmark covering diverse asset classes, \model attains state-of-the-art performance, ranking first across all three evaluation tiers: point-forecast accuracy, cross-sectional asset ranking, and portfolio profitability.

\vspace{0.55em}
\noindent
{\footnotesize
\textbf{Model}\quad\url{https://huggingface.co/LG-AI-Research/EXAONE-Finance-1.0}\\[1pt]
\textbf{Code}\quad\;\url{https://github.com/LGAI-Research/EXAONE-Forecast}
}

%% file: sections/01_introduction.tex
\section{Introduction}
\label{sec:intro}

Time series (TS) forecasting underpins decision-making across domains such as finance~\citep{sezer2020financial,lee2026beyond,lee2026finstar} and energy~\citep{hong2016probabilistic}. To address these forecasting needs, deep learning models, including recurrent~\citep{salinas2020deepar}, convolutional~\citep{bai2018tcn,luo2024moderntcn}, and transformer-based~\citep{zhou2021informer,nie2023patchtst} architectures, capture temporal structure effectively, yet are typically trained for individual datasets and transfer poorly across domains.
To address this limitation, time series foundation models (TSFMs) 
pursue generalizable forecasting through large-scale pretraining~\citep{ansari2024chronos,woo2024unified,das2024decoder,garza2023timegpt,goswami2024moment,lee2026dataset,lee2026not},
achieving 
strong zero-shot performance across diverse TS benchmarks~\citep{godahewa2021monash,aksu2024gifteval}.

Despite their success on general-domain benchmarks, \textit{applying TSFMs to financial forecasting remains challenging} due to three gaps.
\textbf{(i) Cost:} the self-attention backbones used by most TSFMs incur a cost that grows quadratically with the
sequence length and the number of variates, which is expensive for the long,
many-channel panels in finance.
\textbf{(ii) Missingness:} financial series are interrupted by market
closures and trading halts, yet most TSFMs assume contiguous inputs.
\textbf{(iii) Domain mismatch:} pretraining corpora are dominated by
non-financial domains, under-representing financial dynamics such as heavy tails and low
signal-to-noise.

\textbf{Broad view of financial assets.}
Crucially, \emph{financial forecasting} is far broader than stock-price prediction.
Financial 
decision-making spans a heterogeneous universe of assets
with distinct sampling frequencies, scales, and dynamics:
equities, exchange-traded funds (ETFs), foreign exchange (FX), commodities, crypto-assets, fixed-income instruments, and macroeconomic indicators.
These asset classes exhibit diverse characteristics, ranging from high-frequency, heavy-tailed crypto series
to low-frequency, smoothly varying macroeconomic indicators, 
and are often observed
jointly as multi-channel panels (e.g., the open/high/low/close/volume fields of a single security).
A \emph{financial TSFM} should therefore support this 
spectrum of
asset classes,
enabling a single model to address the diverse forecasting needs 
encountered across financial markets.

\begin{wrapfigure}{r}{0.45\textwidth}
\centering
\vspace{-5pt}
\includegraphics[width=0.36\textwidth]{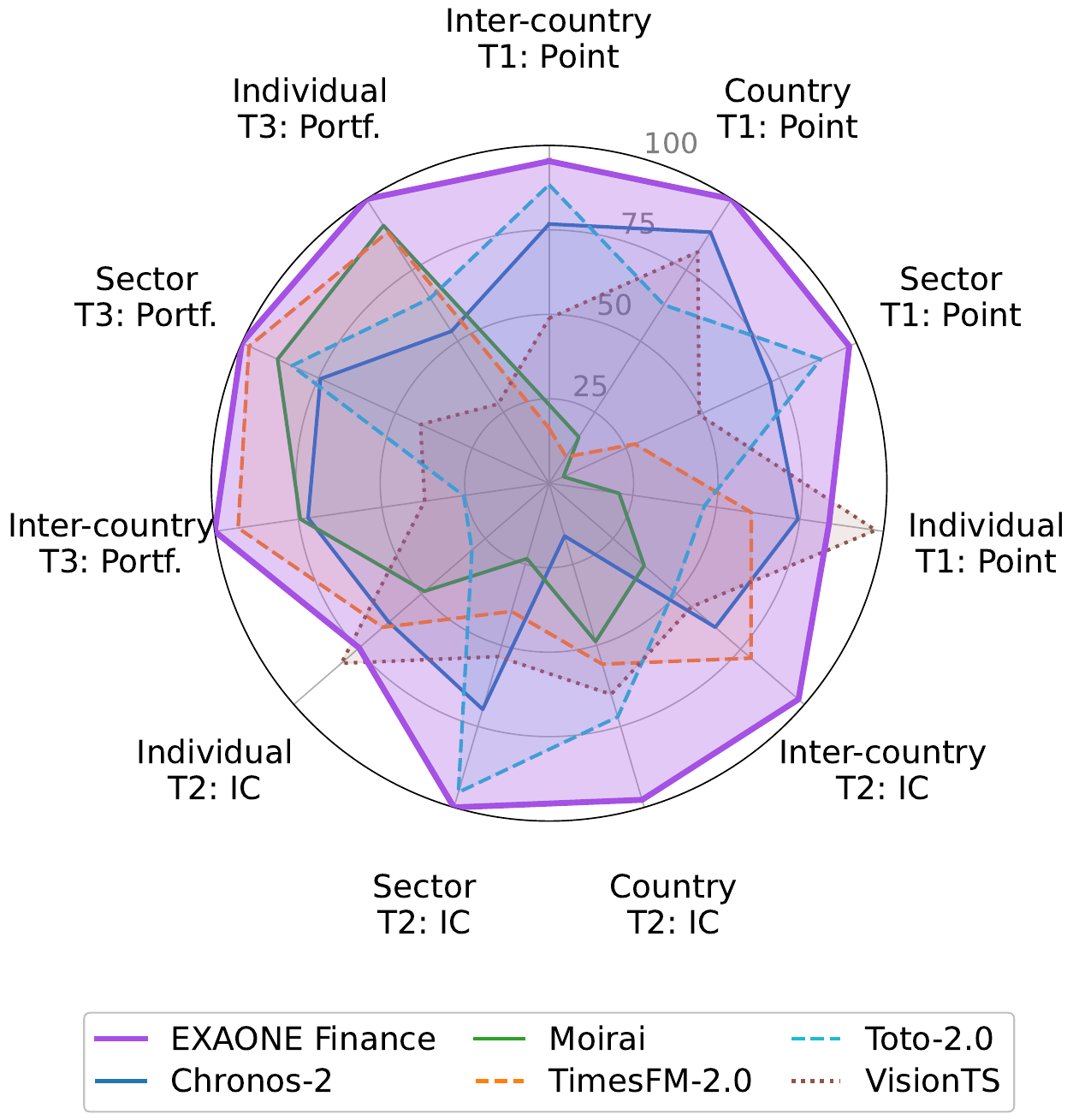}
\vspace{-2pt}
\caption{\textbf{FinVerse benchmark.} \model outperforms five
representative
TSFMs.}
\label{fig:radar}
\vspace{-18pt}
\end{wrapfigure}
To this end, we propose \textbf{\model}, a TSFM tailored to financial forecasting. 
Unlike general-purpose TSFMs, \model is designed to address the unique characteristics of financial TS. It adopts an 1) \textit{attention-free} architecture by replacing self-attention with two linear-time operators, namely a causal 1D convolution for temporal mixing and a group-aware pooling MLP for variate mixing, and is trained with a 2) \textit{masked-context augmentation} that improves robustness to missing observations. 
Furthermore, our method
is pretrained on a 3) \textit{synthetic financial corpus}, generated so that the sampled series carry the properties of financial markets such as heavy tails, volatility clustering, jumps, regime shifts, and cross-asset dependence.
We evaluate it on FinVerse~\citep{lee2026finverse}, a standardized financial benchmark 
that assesses models across three tiers (point accuracy, cross-sectional skill, and portfolio backtesting) over a broad range of financial assets.
Table~\ref{tbl:compare} contrasts \model with representative TSFMs in terms of their mixing architecture and pretraining domain.
The main contributions are 
as follows:

\begin{table}[t]
\centering
\caption{\textbf{Comparison of TSFMs.} \model is the only
\textit{attention-free} convolutional TSFM and the only model in this comparison pretrained on
\textit{financial-domain data}, which we generate synthetically.
(CNN: convolutional neural network, RNN: recurrent
neural network, SSM: state-space model, MLP: multi-layer perceptron)}
\label{tbl:compare}
\vspace{1mm}
\adjustbox{max width=\linewidth}{%
\begin{NiceTabular}{l | c c | c}
\toprule
\multirow{2.5}{*}{\textbf{Models}} & \multicolumn{2}{c}{\textbf{Architecture}} & \textbf{Dataset} \\
\cmidrule(lr){2-3}\cmidrule(lr){4-4}
& \textbf{Temporal mixing} & \textbf{Variate mixing} & \textbf{Financial} \\
\midrule
\makecell[l]{Chronos, Chronos-Bolt, TimesFM, Lag-Llama, TimeGPT, Timer,\\
MOMENT, Sundial, TEMPO, ROSE, Time-MoE, PatchTST-FM,\\
Kairos, YingLong, CleanTS, TabPFN-TS, VisionTS}
  & Attention & -- & \xmark \\
\midrule
\makecell[l]{Chronos-2, Moirai, Moirai-MoE, Toto, UniTS}
  & Attention & Attention & \xmark \\
\midrule
\makecell[l]{TiRex (xLSTM), TempoPFN, FlowState, Reverso}
  & RNN / SSM & -- & \xmark \\
\midrule
TTM (TinyTimeMixer) & MLP & MLP & \xmark \\
\midrule
\rowcolor{LightPurple}
\textbf{\model (Ours)} & \textbf{CNN} & \textbf{MLP (Group pooling)} & \cmark \\
\bottomrule
\end{NiceTabular}}
\vspace{-11pt}
\end{table}

\begin{itemize}[leftmargin=1.2em,itemsep=1pt,topsep=2pt]
\item We propose \textbf{\model}, an attention-free TSFM that mixes time
with \textit{causal convolutions} and variates with a \textit{group-aware pooling
MLP} for linear-time cost (with a patch-based forecasting head), made robust to
missing data by a masked-context augmentation that masks contiguous spans of the input.
\item We pretrain \model on a \textit{synthetic financial corpus}, using a generator
whose sampling process is designed to reproduce the properties of financial series such as heavy
tails, volatility clustering, jumps, regime shifts, and cross-asset dependence. The model
therefore sees no real market data during pretraining.
\item Through 
evaluation on \textbf{FinVerse}, a financial benchmark that spans a broad range of asset
classes and scores models on three tiers (point accuracy, cross-sectional skill, and
portfolio backtesting), \model attains state-of-the-art (SoTA) performance across
financial forecasting tasks. 
As shown in Figure~\ref{fig:radar}, 
it traces a near-maximal envelope across the
scope\,$\times$\,tier grid, ranking in the top percentiles
for nearly every cell.
\end{itemize}

%% file: sections/02_related_work.tex
\section{Related Work}
\label{sec:related}

\subsection{Time Series Foundation Models}
\label{sec:related_tsfm}
Inspired by the success of large language models, 
recent work
pretrains
a single model on large, heterogeneous TS corpora 
for
zero-shot forecasting. Chronos~\citep{ansari2024chronos} tokenizes scalar values and
reuses a T5 encoder--decoder language-model backbone, later distilled into the
faster Chronos-Bolt and extended to multivariate inputs by Chronos-2~\citep{ansari2025chronos2}.
TimesFM~\citep{das2024decoder} adopts a decoder-only, patch-based
transformer, while Moirai~\citep{woo2024unified} uses a masked encoder
with any-variate attention to support arbitrary numbers of covariates, and
Moirai-MoE~\citep{liu2025moiraimoe} sparsifies it with a mixture-of-experts.
Beyond these, the design space 
broadens
through decoder-only
autoregressive models~\citep{rasul2023lagllama,liu2024timer,shi2024timemoe,garza2023timegpt},
encoder-style representation models~\citep{goswami2024moment}, further
foundation models~\citep{cohen2024toto,liu2025sundial,cao2024tempo,gao2024units,wang2024rose},
and even the repurposing of the vision domain for
forecasting~\citep{chen2025visionts}. 
Recently,
Toto-2.0~\citep{khwaja2026toto2} pushes this paradigm into the scaling era with a
family of open models trained at multiple sizes.
A common characteristic of most of these models is their reliance on \emph{self-attention}, whose computational cost grows quadratically with 
the sequence length and the number of variates.
\model departs from
this paradigm 
by removing the attention mechanism,
while
keeping the patch-based, quantile-decoding interface that these models share.
Even the few TSFMs built for finance retain an attention backbone. FinCast~\citep{zhu2025fincast} adopts a channel-independent transformer that models series in isolation, and Delphyne~\citep{ding2025delphyne} relies on quadratic any-variate attention, whereas \model mixes grouped variates jointly in \emph{linear} time.

\subsection{Attention-Free Sequence Models for Time Series}
\label{sec:related_attnfree}
Attention is not the only way to mix information across time. Convolutional
sequence models, from the original Temporal Convolutional Network
(TCN)~\citep{bai2018tcn} to the more recent
ModernTCN~\citep{luo2024moderntcn}, use (dilated) causal convolutions to
obtain large receptive fields at linear cost, and have proven competitive with
transformers on long-horizon forecasting. Purely MLP-based models such as
DLinear~\citep{zeng2023dlinear}, TSMixer~\citep{chen2023tsmixer},
and TiDE~\citep{das2023tide} show that even simple linear or
token-/channel-mixing layers are strong baselines, and TTM
(TinyTimeMixer)~\citep{ekambaram2024ttm} carries this idea into the
foundation-model regime with an MLP-Mixer backbone. State-space models, most
notably Mamba~\citep{gu2024mamba} and its TS
adaptations~\citep{wang2024smamba,lee2024sormamba}, offer
another linear-time alternative with selective recurrence. \model follows this attention-free lineage: it combines a \emph{causal convolution} for temporal mixing
with a \emph{group-aware pooling MLP} for variate mixing. Unlike most prior
attention-free work, trained per-dataset, it is a
\emph{pretrained foundation model}.

\subsection{Large-scale Pretraining Datasets}
\label{sec:related_data}
Pretraining a TSFM requires large and diverse corpora, drawn from real archives,
synthetic generators, or both. On the real side, the Monash
archive~\citep{godahewa2021monash} aggregates dozens of public datasets,
LOTSA~\citep{woo2024unified} scales this to over 27B observations across nine
domains, GIFT-Eval~\citep{aksu2024gifteval} assembles a broad multi-domain corpus,
and Time-300B~\citep{shi2024timemoe} reaches over 300B time points. On the synthetic
side, KernelSynth~\citep{ansari2024chronos} composes Gaussian-process kernels 
to synthesize diverse trend and seasonal
patterns, and CauKer~\citep{xie2025cauker} augments such kernel compositions with
structural causal models for causally coherent inter-series interactions. In
contrast, \model stays synthetic but targets a domain: its generator is built so that
the sampled series reproduce the properties of financial markets
(Section~\ref{sec:model_pretrain}).

%% file: sections/03_model.tex
\section{Problem Definition}
\label{sec:model_problem}
\vspace{-3pt}
\begin{wraptable}{r}{0.40\textwidth}
\vspace{-\intextsep}
\centering
\caption{Notations for TS forecasting.}
\label{tbl:notation}
\footnotesize
\renewcommand{\arraystretch}{0.80}
\adjustbox{max width=\linewidth}{%
\begin{NiceTabular}{c l}
\toprule
\textbf{Symbol} & \textbf{Meaning} \\
\midrule
$L$ & Context (Input) length \\
$H$ & Forecast horizon length \\
$C$ & Number of variates in a group \\
$B$ & Batch size (\# variate-rows) \\
$P$ & Patch size \\
$d$ & Model hidden dimension \\
$\bm{x}^{\mathrm{ctx}}\in\mathbb{R}^{L}$ & Univariate context input \\
$\bm{X}^{\mathrm{ctx}}\in\mathbb{R}^{C\times L}$ & Multivariate context input\\
$\bm{y}\in\mathbb{R}^{H}$ & Ground-truth horizon values \\
$\bm{m}\in\{0,1\}^{L}$ & Observation mask \\
$\bm{g}\in\{1,\dots,G\}^{C}$ & Variate group ids \\
$q_1,\dots,q_Q$ & Target quantile levels \\
$\hat{\bm{Y}}\in\mathbb{R}^{Q\times H}$ & Predicted quantiles \\
\bottomrule
\end{NiceTabular}}
\vspace{-22pt}
\end{wraptable}
\textbf{Time series forecasting.}
Let a single time series be $\bm{x}=(x_1,x_2,\dots)$ observed at a fixed sampling
\emph{frequency} (e.g., daily, weekly, monthly). At a forecast origin $t$, the
model is given a \emph{context window} of the most recent $L$ observations,
$\bm{x}^{\mathrm{ctx}} = (x_{t-L+1},\dots,x_t)\in\mathbb{R}^{L}$, and must predict the
next $H$ values, called the \emph{forecast horizon},
$\bm{y} = (x_{t+1},\dots,x_{t+H})\in\mathbb{R}^{H}$.

\textbf{Multivariate (grouped) series.}
Financial assets are frequently observed as aligned multi-channel panels (e.g.,
the open/high/low/close/volume fields of one security, 
or a group of related assets).
We therefore allow a \emph{group} of $C$ variates sharing common
timestamps,
$\bm{X}^{\mathrm{ctx}}=[\bm{x}^{\mathrm{ctx}}_1;\dots;\bm{x}^{\mathrm{ctx}}_C]\in\mathbb{R}^{C\times L}$,
with a group-id vector $\bm{g}\in\{1,\dots,G\}^{C}$ indicating which variates may
exchange information ($g_i=g_j$ iff variates $i,j$ belong to the same group).
A univariate series is the special case $C=1$.

\textbf{Missingness.}
Financial series contain missing spans 
(e.g., market closures, trading halts, and unreported periods).
We represent observability with an observation mask
$\bm{m}\in\{0,1\}^{L}$, where $m_\tau=0$ marks a missing/padded entry. The model
must handle such missing spans and forecast through them without imputation.

\textbf{Probabilistic objective.}
Rather than a single point estimate, we produce a set of $Q$ quantile forecasts.
Given target quantile levels
$q_1,\dots,q_Q\in(0,1)$, the model
$f_{\bm{\theta}}$ maps the (masked) context to
\begin{equation}
\hat{\bm{Y}} = f_{\bm{\theta}}\big(\bm{X}^{\mathrm{ctx}},\bm{m},\bm{g}\big)
\in\mathbb{R}^{Q\times H},
\qquad
\hat{y}_{q,h}\approx \mathrm{Quantile}_q\!\left(x_{t+h}\right),
\label{eq:objective}
\end{equation}
where $\hat{y}_{q,h}$ estimates the $q$-th quantile of the $h$-step-ahead value.
Table~\ref{tbl:notation} summarizes the notation.

\section{\model}
\label{sec:model}

\begin{figure*}[t]
\centering
\adjustbox{max width=\linewidth}{\includegraphics{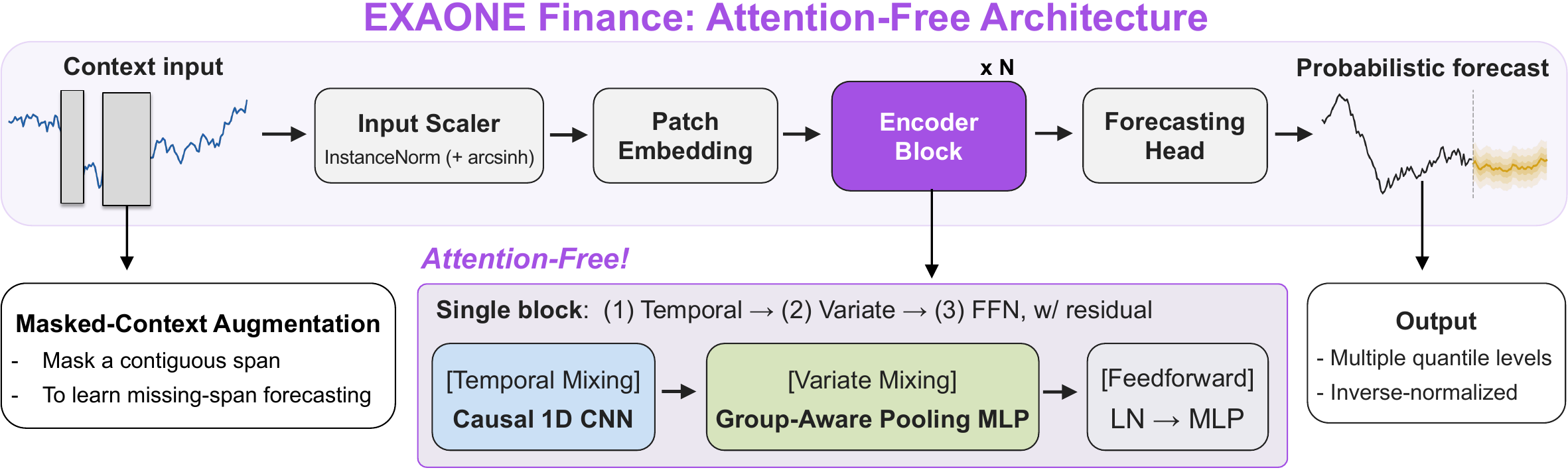}}
\caption{
\textbf{Architecture of \model.} 
The context series is
instance-normalized,
split into patches, and embedded, then processed
by $N$ \emph{attention-free} encoder blocks before a forecasting head produces a
probabilistic forecast. Each encoder block mixes information with two linear-time
operators: a \emph{causal 1D convolution} for temporal mixing and a
\emph{group-aware pooling MLP} for variate mixing, followed by a feed-forward
network.
During training,
a \emph{masked-context augmentation} marks a contiguous input span as
known-missing, so the model learns to forecast through the missing spans 
that occur at inference.}
\label{fig:arch}
\end{figure*}

\subsection{Model Overview}
\label{sec:model_overview}
\model is an attention-free TSFM that replaces self-attention with two simple yet effective linear-time mixers, where each encoder block combines (i) a \textit{causal 1D convolution} over the temporal axis (Section~\ref{sec:model_temporal}) and (ii) a \textit{group-aware pooling MLP} over the variate axis (Section~\ref{sec:model_variate}), followed by a standard feed-forward network (FFN).
This makes the per-block cost linear in both the sequence length $L$ and
the number of variates $C$.
\textcolor{black}{Missing observations are handled natively using an observation-mask channel that flags missing or padded entries in each patch token (Section~\ref{sec:model_input}). Instance normalization (Eq.~\ref{eq:instnorm}) and the training loss (Section~\ref{sec:model_head}) ignore these positions, allowing the model to process and forecast through missing spans without a separate imputation step.}
The end-to-end pipeline is summarized as follows:
\begin{equation}
\bm{x}^{\mathrm{ctx}}
\;\xrightarrow{\text{InstanceNorm}}\;
\hat{\bm{x}}
\;\xrightarrow{\text{Patch}}\;
\bm{P}
\;\xrightarrow{\text{Embed}}\;
\bm{H}^{(0)}
\;\xrightarrow{\;N\times\text{Encoder}\;}\;
\bm{H}^{(N)}
\;\xrightarrow{\text{Head}}\;
\hat{\bm{Y}} .
\label{eq:pipeline}
\end{equation}
We detail the input pipeline (Section~\ref{sec:model_input}),
the two mixing operators (Sections~\ref{sec:model_temporal}--\ref{sec:model_variate}),
a masked-context training augmentation (Section~\ref{sec:model_mask}),
and the forecasting head (Section~\ref{sec:model_head}).
The overall framework of \model is illustrated in Figure~\ref{fig:arch}.

\subsection{Input Pipeline}
\label{sec:model_input}
\textbf{Instance normalization.}
Each context is standardized along the time axis using statistics: 
\begin{equation}
\hat{\bm{x}} = \frac{\bm{x}^{\mathrm{ctx}}-\mu}{\sigma},
\qquad
\mu=\operatorname{mean}(\bm{x}^{\mathrm{ctx}}),\;\;
\sigma=\operatorname{std}(\bm{x}^{\mathrm{ctx}}),
\label{eq:instnorm}
\end{equation}
where $\mu$ and $\sigma$ are computed over the \emph{observed} (non-missing) entries only
and used
to restore the predictions to the original scale.
To accommodate the heavy-tailed nature of financial series, an area hyperbolic sine
($\operatorname{arcsinh}$) transform is then applied to the standardized context:
\begin{equation}
\tilde{\bm{x}} \;=\; \operatorname{arcsinh}(\hat{\bm{x}})
\;=\; \log\!\Big(\hat{\bm{x}} + \sqrt{\hat{\bm{x}}^{2}+1}\,\Big).
\label{eq:arcsinh}
\end{equation}
This transform behaves linearly near zero ($\operatorname{arcsinh}(u)\!\approx\!u$) and
logarithmically in the tails ($\operatorname{arcsinh}(u)\!\approx\!\operatorname{sgn}(u)\,\log 2|u|$),
compressing extreme values while preserving sign. Unlike a log transform, it remains
defined for negative and zero inputs. It reduces to the identity ($\tilde{\bm{x}}\!=\!\hat{\bm{x}}$)
when disabled, and is inverted at inference by $\hat{\bm{x}}=\sinh(\tilde{\bm{x}})$ before
de-standardization.

\textbf{Patching.}
The transformed context $\tilde{\bm{x}}$ is unfolded into non-overlapping patches of size $P$
(stride $P$). When the length is not a multiple of $P$, the sequence is
left-padded with missing markers so that the most recent observations remain
patch-aligned. Each patch token concatenates three components,
\begin{equation}
\bm{p}_\tau = \big[\, \bm{e}_\tau \;\|\; \tilde{\bm{x}}_\tau \;\|\; \bm{m}_\tau \,\big]\in\mathbb{R}^{3P},
\label{eq:patch}
\end{equation}
a time encoding $\bm{e}_\tau$ (each position's index relative to the forecast origin, divided by a fixed scale), the patched values $\tilde{\bm{x}}_\tau$, and an
observation mask $\bm{m}_\tau$ that marks padded/missing entries. Each token is then
embedded into the model dimension by a residual MLP block: it takes the
$3P$-dimensional token, passes it through a hidden layer of width
$d_{\mathrm{ff}}$, and outputs a $d$-dimensional representation. Stacking the
embedded tokens gives the encoder input $\bm{H}^{(0)}\in\mathbb{R}^{T\times d}$,
where $T$ is the number of patches. An optional learnable \emph{register} token is then
prepended to this sequence.

\subsection{Temporal Mixing}
\label{sec:model_temporal}
\textbf{Causal convolution block.}
Self-attention along time is replaced by a stack of $n$ causal 1D convolutions.
Given block input $\bm{H}\in\mathbb{R}^{T\times d}$, we layer-normalize, transpose
to channel-first, and apply each convolution with a left-only pad of
$k\!-\!1$ so that token $\tau$ depends solely on tokens $\le\tau$
\textcolor{black}{(ensuring causality)}:
\begin{equation}
\bm{U} = \mathrm{GELU}\!\big(\mathrm{Conv1d}_{k}(\mathrm{LeftPad}_{k-1}(\cdot))\big),
\quad \text{(repeated $n$ times)},
\label{eq:cnn_stack}
\end{equation}
where the first convolution maps $d\!\to\!c$ channels and the rest operate in $c$.
A final $1\!\times\!1$ convolution restores the model width $c\!\to\!d$, and a
residual connection is added:
\begin{equation}
\bm{H}' = \bm{H} + \mathrm{Dropout}\!\big(\mathrm{Conv1d}_{1\times1}(\bm{U})\big).
\label{eq:cnn_res}
\end{equation}

\subsection{Variate Mixing}
\label{sec:model_variate}
\begin{figure}[t]
\centering
\adjustbox{max width=\linewidth}{\includegraphics{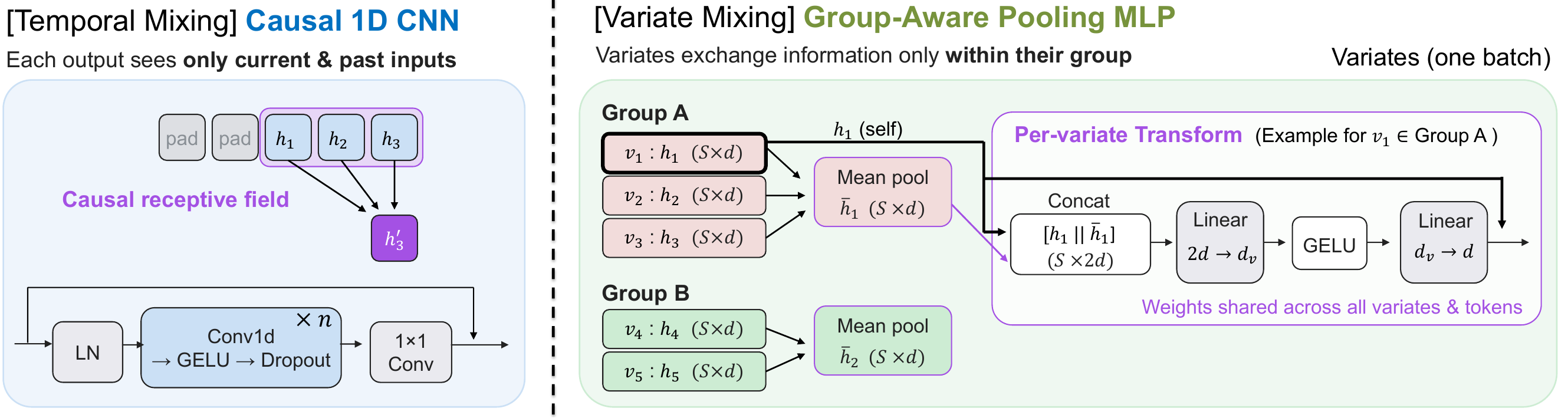}}
\caption{\textbf{The two attention-free mixing operators.} Each encoder block mixes
information along two axes. [Left] \emph{Temporal mixing}  applies a causal 1D
convolution, 
so that each position attends only to current and past
inputs. [Right] \emph{Variate mixing} is a group-aware pooling MLP,
where variates exchange information only within their group, and each variate
$\bm{h}_i$ is concatenated with its group mean $\bar{\bm{h}}_i$ and passed through a
shared MLP with a residual connection.}
\label{fig:variate_pool}
\end{figure}

\textbf{Group-aware pooling MLP.}
Information is exchanged across variates only \emph{within the same group},
so that unrelated series are never
mixed. Let $g_i$ denote the group id of variate $i$ and define the group mask
$\Gamma_{ij}=\mathbb{1}[g_i=g_j]$. Each variate is augmented with its group-mean
representation,
\begin{equation}
\bar{\bm{h}}_i
= \frac{\sum_j \Gamma_{ij}\,\bm{h}_j}{\sum_j \Gamma_{ij}},
\qquad
\bm{h}'_i
= \bm{h}_i + \mathrm{Dropout}\!\big(\mathrm{MLP}([\,\bm{h}_i \,\|\, \bar{\bm{h}}_i\,])\big),
\label{eq:variate}
\end{equation}
where the MLP projects $2d\!\to\!d_v\!\to\!d$ with a GELU nonlinearity.
Since each variate is mixed only through its group mean, the output does not
depend on the order of variates within a group. When a group contains only a
single variate, no cross-variate information is available, and the operator
reduces to a per-variate transformation. After the two mixers, a standard
FFN is applied, that is, layer normalization (LN) followed by a two-layer MLP with hidden dimension $d_{\mathrm{ff}}$, with a residual connection. Both mixers are shown in Figure~\ref{fig:variate_pool}.

\subsection{Masked-Context Augmentation}
\label{sec:model_mask}
\textbf{Robustness to missingness.}
Financial series are routinely interrupted by market closures, trading halts,
and unreported periods. To make the encoder robust to such missing spans, during
training we randomly mark a single contiguous span of each context as
\emph{missing} before normalization,
\begin{equation}
\tilde{x}_\tau =
\begin{cases}
\text{missing}, & \tau\in[s,\,s+w)\\
x_\tau, & \text{otherwise,}
\end{cases}
\qquad w \sim \mathcal{U}\{1,\dots,\rho L\},
\label{eq:mask}
\end{equation}
with masking fraction $\rho=0.25$ and a uniformly sampled start $s$.
Since the normalization statistics, 
the observation mask,
and the loss
all treat missing
entries consistently as ``known-missing,'' the model learns to forecast through missing spans. Note that the augmentation is applied
\emph{only at training time}.

\textbf{Comparison with masking in prior works.}
Most existing masking schemes in TS are
reconstruction
objectives~\citep{zerveas2021transformer,dong2023simmtm,lee2024pits}, in which the masked span is itself the \emph{prediction target}, i.e.,
masking is applied to the \emph{output} to be reconstructed. In contrast, our augmentation masks part of the \textit{input} rather than the target: the masked span is marked known-missing and excluded from the loss, so that the model becomes robust to the missing observations it encounters at inference, which are common in financial markets (e.g., market closures and trading halts).

\subsection{Forecasting Head}
\label{sec:model_head}
\textbf{Quantile decoding.}
After $N$ encoder blocks and a final LN, the last
$\lceil H/P\rceil$ tokens are decoded by the forecasting head (a residual MLP) into
$Q\!\cdot\!P$ outputs and rearranged into quantile trajectories
$\hat{\bm{Y}}\in\mathbb{R}^{Q\times H}$, which are mapped back to the original
scale by inverting Eq.~\ref{eq:instnorm}. We predict a set of $Q$ quantile levels
and train with the pinball (quantile)
loss, evaluated only at finite target positions via a mask $\bm{m}^{y}$:
\begin{equation}
\mathcal{L}
= \frac{1}{|\bm{m}^{y}|}\sum_{q}\sum_{h} m^{y}_{h}\,
\max\!\big(q\,(y_h-\hat{y}_{q,h}),\;(q-1)(y_h-\hat{y}_{q,h})\big).
\label{eq:pinball}
\end{equation}
Masking the loss lets multivariate groups feed all channels as \emph{inputs}
while scoring only the designated target channel(s), matching the evaluation
protocol.

\subsection{Configuration, Inference, and Complexity}
\label{sec:model_config}
\begin{wraptable}{r}{0.42\textwidth}
\vspace{-\intextsep}\vspace{-5pt}
\centering
\caption{Architecture configuration.}
\label{tbl:arch}
\vspace{-3pt}
\adjustbox{max width=\linewidth}{%
\begin{NiceTabular}{l r}
\toprule
\textbf{Hyperparameter} & \textbf{Value} \\
\midrule
\rowcolor{Band}\textbf{Encoder} & \\
\quad Hidden dimension ($d$)               & 1024 \\
\quad Feed-forward dimension ($d_{\mathrm{ff}}$) & 4096 \\
\quad Number of blocks ($N$)               & 12 \\
\quad Dropout                              & 0.1 \\
\rowcolor{Band}\textbf{Encoder block} & \\
\quad\textit{Temporal CNN} & \\
\quad\quad Channels ($c$)                  & 512 \\
\quad\quad Kernel size ($k$)               & 7 \\
\quad\quad Conv layers ($n$)               & 2 \\
\quad\textit{Variate MLP} & \\
\quad\quad Hidden dimension ($d_v$)        & 512 \\
\quad\quad Max groups ($G$)                & 16 \\
\rowcolor{Band}\textbf{Patch \& head} & \\
\quad Patch size ($P$)                     & 16 \\
\quad Max input length ($L$)               & 512 \\
\quad Quantile levels ($Q$)                & 21 \\
\bottomrule
\end{NiceTabular}}
\vspace{-20pt}
\end{wraptable}
\textbf{Model configuration.}
Table~\ref{tbl:arch} summarizes the hyperparameters of the released model. The encoder
stacks $N{=}12$ blocks of width $d{=}1024$, each pairing a temporal CNN ($c{=}512$
channels, kernel $k{=}7$, $n{=}2$ layers) with a group-aware variate MLP (hidden size
$d_v{=}512$, up to $G{=}16$ groups). The forecasting head reads out patches of size
$P{=}16$ as $Q{=}21$ quantile levels.

\textbf{Single forward pass.}
At inference, \model decodes the entire horizon in \emph{one} forward pass
rather than autoregressively. For a requested horizon $H$, the model appends
$\lceil H/P\rceil$ future tokens to the context, runs the $N$ encoder blocks once,
and reads out all $Q$ quantiles for every horizon step simultaneously
(Section~\ref{sec:model_head}). No step-by-step roll-out and no key/value cache are
needed, as there is no attention to cache. Arbitrary horizons are supported by varying
the number of output tokens, and the prediction is mapped back to the original scale
by inverting Eq.~\ref{eq:instnorm}. Missing or padded entries in the context are
handled by the same observation mask used in training, so no special preprocessing
(e.g., imputation) is required at inference.
This scheme is identical during pretraining: the
$\lceil H/P\rceil$ future tokens are all appended up front and predicted jointly in a
single pass.

\textbf{Computational complexity.}
Both mixers are linear in their respective axes. A causal convolution block costs
$\mathcal{O}(L\,d\,c\,k)$ for sequence length $L$, and the group-aware pooling MLP
costs $\mathcal{O}(C\,d)$ for $C$ variates: each group mean is computed once by averaging within each group (written with a group mask in Eq.~\ref{eq:variate} only for clarity), not by a
pairwise interaction. This contrasts with the $\mathcal{O}(L^2 d)$ time-axis
attention and $\mathcal{O}(C^2 d)$ variate-axis attention in TSFMs that adopt
self-attention, making \model linear in both the sequence length and the
number of variates.

%% file: sections/04_experiments.tex
\section{Experiments}
\label{sec:exp}
We evaluate \model on \textbf{FinVerse}, a standardized, unified
benchmark covering a broad range of financial assets. We first describe the pretraining dataset
(Section~\ref{sec:model_pretrain}) and the evaluation protocol, namely the
three scoring tiers and their metrics (Section~\ref{sec:exp_protocol}), then the baselines
and aggregation (Section~\ref{sec:exp_setup}), the main results
(Section~\ref{sec:exp_main}), and an analysis of performance broken down by asset scope (Section~\ref{sec:exp_asset}).

\subsection{Pretraining Dataset}
\label{sec:model_pretrain}
\textbf{Data mixture.}
We pretrain \model on a synthetic corpus designed around the nature of financial data. It
combines a domain-agnostic source, KernelSynth~\citep{ansari2024chronos}, with a
\emph{synthetic financial} source whose sampling process we design so that the generated
series carry the statistical regularities of financial markets. \model therefore sees
\textit{no real market data at any point during pretraining}, yet transfers to real assets
at evaluation.

\begin{figure}[t]
\centering
\includegraphics[width=\linewidth]{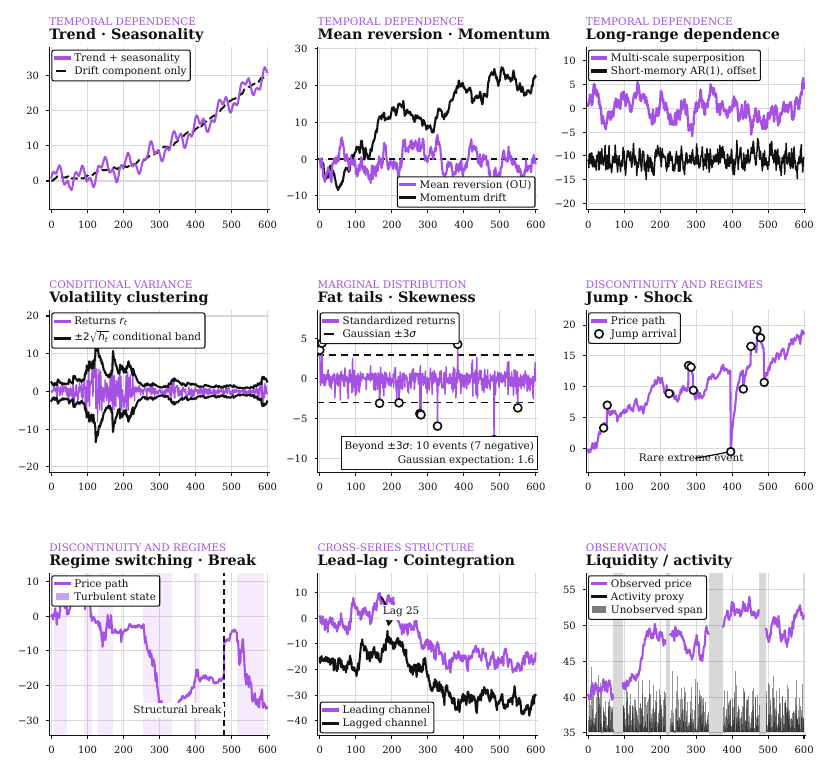}
\caption{\textbf{Properties of the synthetic pretraining corpus.} Each panel simulates one
property of Table~\ref{tbl:stylized} with the generator component that induces it. The purple
curve is the series that shows the property. The black curve, band, or markers are there for
comparison, and what they mean changes from panel to panel, so each panel has its own legend.}
\label{fig:stylized}
\end{figure}

\textbf{Synthetic financial datasets.}
Table~\ref{tbl:stylized} lists the properties we ask the generator to reproduce, grouped
into six families, together with the mechanism that induces each. We sample a series as a
superposition of these components: a trend and cyclical part sets the low-frequency
shape, an ARMA / Ornstein--Uhlenbeck part sets the short-horizon dependence, a
stochastic-volatility part with heavy-tailed, skewed innovations sets the conditional
distribution, a jump and regime part introduces discontinuities and parameter changes,
and a factor part ties channels together so that a group shares common shocks with
lead--lag and cointegration structure. We switch every property on with a randomly
sampled intensity. The corpus therefore spans series close to Gaussian random walks as
well as series dominated by clustered volatility and abrupt regime changes.
Figure~\ref{fig:stylized} shows a representative property of each family as it appears in a
sampled series, next to the reference behaviour that the corresponding component departs from.

\begin{table}[t]
\centering
\caption{\textbf{Financial time series properties and their generating mechanisms.} Each
property is induced by an explicit component of the sampling process. Properties are
switched on independently with sampled intensities.}
\label{tbl:stylized}
\vspace{1mm}
\adjustbox{max width=\linewidth}{%
\begin{NiceTabular}{l l l}
\toprule
\textbf{Family} & \textbf{Property} & \textbf{Generating mechanism} \\
\midrule
\multirow{7}{*}{\makecell[l]{\textbf{Temporal}\\\textbf{dependence}}}
 & Trend                     & Deterministic drift plus a random-walk drift component \\
 & Seasonality / cyclicality & Fourier components at calendar periods \\
 & Autocorrelation           & ARMA innovations \\
 & Mean reversion            & Ornstein--Uhlenbeck component \\
 & Momentum                  & Positive short-horizon dependence in the drift \\
 & Long-range dependence     & Fractional Gaussian noise with sampled Hurst exponent \\
 & Multi-scale dynamics      & Superposition of components at several time scales \\
\midrule
\multirow{3}{*}{\makecell[l]{\textbf{Conditional}\\\textbf{variance}}}
 & Volatility                & Sampled base volatility level \\
 & Volatility clustering     & GARCH-type conditional variance \\
 & Heteroskedasticity        & Time-varying variance schedules \\
\midrule
\multirow{3}{*}{\makecell[l]{\textbf{Marginal}\\\textbf{distribution}}}
 & Fat tails                 & Student-$t$ innovations with sampled degrees of freedom \\
 & Skewness                  & Skewed innovations and a leverage effect \\
 & Non-normality             & Mixture-of-normals innovation distribution \\
\midrule
\multirow{4}{*}{\makecell[l]{\textbf{Discontinuity}\\\textbf{and regimes}}}
 & Jump / discontinuity      & Compound-Poisson jump component \\
 & Shock / extreme event     & Rare large-amplitude shocks with decaying response \\
 & Regime switching          & Markov-switching parameters \\
 & Structural break          & Change-points in the parameter path \\
\midrule
\multirow{4}{*}{\makecell[l]{\textbf{Cross-series}\\\textbf{structure}}}
 & Cross-asset correlation   & Common factors with sampled loadings \\
 & Time-varying correlation  & Dynamic conditional correlation of the factors \\
 & Lead--lag relationship    & Lagged factor loadings across channels \\
 & Cointegration             & Shared stochastic trend with a stationary spread \\
\midrule
\textbf{Observation}
 & Liquidity / activity      & Intermittent observation and activity-driven volume proxies \\
\bottomrule
\end{NiceTabular}}
\end{table}

\begin{wraptable}{r}{0.40\textwidth}
\vspace{\dimexpr-\intextsep-2pt\relax}
\centering
\caption{Frequency-aware windows.
}
\label{tbl:window}
\vspace{1pt}
\adjustbox{max width=\linewidth}{%
\begin{NiceTabular}{l c l}
\toprule
\textbf{Frequency} & \textbf{$L$} & \textbf{$H$} \\
\midrule
Daily     & 260 & $\{5,\,20,\,65,\,130,\,260\}$ \\
Weekly    & 52  & $\{1,\,4,\,12,\,26,\,52\}$ \\
Monthly   & 12  & $\{1,\,3,\,6,\,12,\,24\}$ \\
Quarterly & 4   & $\{1,\,2,\,4,\,8\}$ \\
Annual    & 10  & $\{1,\,2,\,5,\,10\}$ \\
\bottomrule
\end{NiceTabular}}
\vspace{-10pt}
\end{wraptable}
\textbf{Frequency-aware windowing.}
Synthetic financial series are generated at the frequencies of
Table~\ref{tbl:window} and windowed with the geometry listed there, which is the same
geometry used at evaluation. KernelSynth series carry no frequency, so they instead use a
variable context length up to $L$ and a horizon sampled uniformly up to $H_{\max}$,
broadening the pretraining distribution. In this run every series is
treated as an individual channel. The group-aware mixer
(Section~\ref{sec:model_variate}) natively supports richer cross-series grouping through
the data pipeline, such as the OHLCV fields of a single security.

\subsection{Evaluation Protocol: Tiers and Metrics}
\label{sec:exp_protocol}

\begin{table}[t]
\centering
\caption{\textbf{Tier metrics.} This table lists the metrics used by the three benchmark
tiers.
}
\label{tbl:metrics}
\vspace{1mm}
\adjustbox{max width=\linewidth}{%
\begin{NiceTabular}{l | l l}
\toprule
\textbf{Tier} & \textbf{Metric} & \textbf{Meaning} \\
\midrule
\multirow{3}{*}{\makecell[l]{\textbf{Tier 1}\\ Point accuracy}}
 & MASE\,$\downarrow$ & Mean absolute error scaled by an in-context na\"ive forecast \\
 & HR / HR$'$ / HR$''$\,$\uparrow$ & Directional accuracy (Sign of change, stricter variants) \\
 & MASE (Diff / raw)\,$\downarrow$ & Differenced and raw-value MASE variants \\
\midrule
\makecell[l]{\textbf{Tier 2}\\ Cross-sectional IC}
 & IC\,$\uparrow$ & Cross-sectional Spearman correlation of predicted vs.\ realized change \\
\midrule
\multirow{4}{*}{\makecell[l]{\textbf{Tier 3}\\ Portfolio backtest}}
 & Return\,$\uparrow$ & Annualized return of the top-$K\%$ long portfolio \\
 & Sharpe\,$\uparrow$ & Annualized return divided by volatility \\
 & Volatility\,$\downarrow$ & Annualized volatility of the strategy \\
 & MDD\,$\uparrow$ & Maximum peak-to-trough loss (Closer to 0 is better) \\
\bottomrule
\end{NiceTabular}}
\end{table}

\textbf{Why finance-specific metrics?}
Standard forecasting metrics such as the mean absolute error (MAE) measure average pointwise
error and are scale-dependent. On a heterogeneous financial panel, a few high-variance series dominate the average error and reward predicting the raw level rather than the actionable signal.
Financial decisions, however, hinge on different quantities: the \emph{direction}
of the next move, the \emph{relative ordering} of assets at a given time, and the
\emph{risk-adjusted return} of a strategy built on the forecasts. FinVerse is
therefore designed around these finance-specific desiderata and organizes
evaluation into three tiers.
Tier~1 uses scale-free, change-based accuracy. Tier~2 measures cross-sectional skill.
Tier~3 measures realized economic value through a portfolio backtest.

\textbf{Change-based targets.}
Tiers 1--2 are computed on a grid of a \emph{change period} (how far back the change
is measured)
and a \emph{forecast position}
(how far ahead of the origin it is evaluated). For an origin $t$, denoting the change
period by $p$ and the forecast position by $h$, the \emph{predicted} and
\emph{realized} changes are
\begin{equation}
\Delta^{\mathrm{pred}} = \frac{\hat{y}_{\,t+h}}{x_{t-p}}-1,
\qquad
\Delta^{\mathrm{act}} = \frac{x_{t+h}}{x_{t-p}}-1,
\label{eq:change}
\end{equation}
where $x$ denotes the ground-truth series, so the base value $x_{t-p}$ is measured a
period $p$ before the origin and both changes run to the same future point $t+h$.
These changes underlie all three tiers.

The three tiers, whose metrics are summarized in Table~\ref{tbl:metrics}, are defined as follows.
\begin{itemize}
\item \textbf{Tier 1 (Point accuracy).}
Per series, we report (i) \textbf{MASE}, the mean absolute error scaled by an
in-context naïve forecast, in both a relative-change and an absolute-difference
variant, plus a raw all-position variant, and (ii) the
\textbf{hit rate (HR)}, the directional accuracy
$\mathrm{HR}=\operatorname{mean}\!\big[\operatorname{sign}(\Delta^{\mathrm{pred}})=\operatorname{sign}(\Delta^{\mathrm{act}})\big]$,
together with stricter variants $\mathrm{HR}'$/$\mathrm{HR}''$ that additionally
require the sign of the first/second difference
to match.

\item \textbf{Tier 2 (Cross-sectional Information Coefficient, IC).}
For each origin $t$, we rank assets within a subgroup by predicted change and compute
the Spearman rank correlation against the realized change,
\begin{equation}
\mathrm{IC}(t)=\operatorname{SpearmanCorr}\big(\{\Delta^{\mathrm{pred}}_s(t)\}_{s},\,\{\Delta^{\mathrm{act}}_s(t)\}_{s}\big),
\qquad
\mathrm{IC}=\operatorname{mean}_t \mathrm{IC}(t).
\label{eq:ic}
\end{equation}
IC measures whether a model correctly orders assets cross-sectionally, which is the
quantity that matters for relative-value strategies.

\item \textbf{Tier 3 (Portfolio backtest).}
We form a top-$K\%$, equal-weight, long-only portfolio by selecting the assets with
the highest predicted change, rebalanced at horizon $h$, in a multi-start overlapping
backtest. We report the annualized \textbf{Return}, \textbf{Sharpe} ratio,
\textbf{Volatility}, and \textbf{Maximum Drawdown (MDD)}. This tier turns forecasts
into an executable strategy and scores the economic value of the predictions.
\end{itemize}

\textbf{Overall ranking.}
We rank models \emph{per cell}: for every (tier, spec, metric) triple, all $M$ models
are ranked from $1$ (best) to $M$. We aggregate these ranks with the \emph{geometric
mean} rather than the arithmetic mean because it is scale-free and rewards
\emph{consistent} performance across heterogeneous cells, since a single very poor rank is
penalized proportionally rather than absolutely. A model's score on a tier is
\begin{equation}
\bar{r} = \Big(\textstyle\prod_{c} r_c\Big)^{1/|\mathcal{C}|}
= \exp\!\Big(\tfrac{1}{|\mathcal{C}|}\textstyle\sum_{c} \ln r_c\Big),
\label{eq:geomean}
\end{equation}
where $r_c$ is the model's rank on cell $c$ and $\mathcal{C}$ is that tier's set of cells. The
headline \emph{rank-sum} is the sum of the three per-tier placements (lower is better).

\begin{table*}[t]
\centering
\caption{\textbf{Main benchmark results on FinVerse.} Each tier gives the model's placement (\#) and its \emph{Avg.\ Rank} (The geometric mean of per-cell ranks, lower is better). \textbf{Rank Sum} is the sum of the three placements, and the 44 models are ordered from the lowest rank sum to the highest. \model is \#1 in all three tiers, for a rank sum of 3.
}
\label{tbl:mainB}
\vspace{1mm}
\small
\adjustbox{max width=\linewidth}{%
\begin{NiceTabular}{c l | c c | c c | c c | c}
\toprule
\multirow{2.5}{*}{\textbf{\#}} & \multirow{2.5}{*}{\textbf{Model}} & \multicolumn{2}{c|}{\textbf{Tier 1}} & \multicolumn{2}{c|}{\textbf{Tier 2}} & \multicolumn{2}{c|}{\textbf{Tier 3}} & \multirow{2.5}{*}{\textbf{Rank Sum}} \\
\cmidrule(lr){3-4}\cmidrule(lr){5-6}\cmidrule(lr){7-8}
 & & \# & Avg.\,Rank & \# & Avg.\,Rank & \# & Avg.\,Rank & \\
\midrule
\rowcolor{LightPurple}
\textbf{1} & \textbf{\model} & \textbf{1} & \textbf{6.53} & \textbf{1} & \textbf{7.45} & \textbf{1} & \textbf{7.43} & \textbf{3} \\
2 & Chronos-2 (Synthetic) & 9 & 11.68 & 3 & 8.74 & 2 & 9.90 & 14 \\
3 & Reverso (Small) & 6 & 9.60 & 2 & 8.64 & 6 & 13.04 & 14 \\
4 & TiRex-1.1 & 2 & 6.98 & 4 & 9.71 & 14 & 15.76 & 20 \\
5 & Chronos-2 & 8 & 10.14 & 13 & 16.12 & 9 & 15.17 & 30 \\
6 & TimesFM-2.5-200M & 7 & 9.96 & 5 & 10.66 & 20 & 17.10 & 32 \\
7 & Chronos-Bolt (Base) & 11 & 12.88 & 15 & 16.48 & 7 & 14.54 & 33 \\
8 & Chronos-Bolt (Small) & 12 & 12.94 & 12 & 15.85 & 16 & 16.10 & 40 \\
9 & T0-Alpha & 13 & 13.47 & 25 & 18.73 & 3 & 11.11 & 41 \\
10 & TiRex-2 (Pretrain) & 3 & 7.48 & 6 & 11.56 & 34 & 20.12 & 43 \\
11 & TiRex-2 (Zero-shot) & 4 & 8.37 & 7 & 11.67 & 33 & 20.06 & 44 \\
12 & Toto-2.0-2.5B & 18 & 15.74 & 17 & 16.78 & 10 & 15.19 & 45 \\
13 & VisionTS & 10 & 12.05 & 8 & 13.00 & 29 & 19.31 & 47 \\
14 & PatchFM & 5 & 9.08 & 14 & 16.15 & 30 & 19.56 & 49 \\
15 & Toto-2.0-4M & 15 & 14.44 & 24 & 18.21 & 13 & 15.66 & 52 \\
16 & TimesFM-2.0-500M & 39 & 29.32 & 10 & 14.38 & 4 & 11.12 & 53 \\
17 & TempoPFN & 14 & 14.11 & 18 & 16.95 & 27 & 18.76 & 59 \\
18 & Kairos-50M & 20 & 16.04 & 29 & 20.21 & 11 & 15.26 & 60 \\
19 & Toto-2.0-1B & 16 & 15.00 & 23 & 17.96 & 22 & 17.74 & 61 \\
20 & FlowState (Granite R1) & 21 & 17.00 & 27 & 19.12 & 15 & 16.09 & 63 \\
21 & TimesFM-1.0-200M & 22 & 17.06 & 16 & 16.58 & 26 & 18.55 & 64 \\
22 & Super-Linear & 24 & 18.11 & 11 & 15.15 & 31 & 19.58 & 66 \\
23 & Toto-2.0-22M & 19 & 15.90 & 26 & 18.97 & 21 & 17.30 & 66 \\
24 & Kairos-10M & 28 & 20.85 & 28 & 19.80 & 12 & 15.61 & 68 \\
25 & FlowState & 23 & 17.49 & 30 & 20.22 & 17 & 16.34 & 70 \\
26 & Toto-2.0-313M & 17 & 15.34 & 31 & 20.55 & 23 & 17.96 & 71 \\
27 & Timer-S1 & 25 & 18.45 & 9 & 13.50 & 39 & 22.81 & 73 \\
28 & Moirai-1.1-R (Large) & 38 & 27.84 & 35 & 22.17 & 5 & 13.00 & 78 \\
29 & Toto-Open-Base-1.0 & 27 & 19.94 & 33 & 21.36 & 19 & 17.07 & 79 \\
30 & Moirai-1.1-R (Base) & 37 & 27.45 & 38 & 24.21 & 8 & 14.57 & 83 \\
31 & Moirai-1.1-R (Small) & 33 & 25.62 & 32 & 21.15 & 18 & 16.87 & 83 \\
32 & PatchTST-FM-R1 & 34 & 26.30 & 22 & 17.92 & 28 & 19.03 & 84 \\
33 & Sundial-Base-128M & 26 & 19.73 & 20 & 17.66 & 41 & 23.48 & 87 \\
34 & PatchTST-FM (Granite R1) & 40 & 29.35 & 19 & 17.34 & 32 & 20.05 & 91 \\
35 & Kairos-23M & 29 & 21.69 & 40 & 24.41 & 24 & 18.35 & 93 \\
36 & TTM-R3 & 35 & 27.10 & 21 & 17.73 & 43 & 24.22 & 99 \\
37 & Moirai-2.0-R (Small) & 31 & 23.87 & 34 & 21.91 & 36 & 21.39 & 101 \\
38 & TTM-R1 & 32 & 24.77 & 44 & 27.22 & 25 & 18.50 & 101 \\
39 & CleanTS-65M & 36 & 27.13 & 36 & 22.70 & 35 & 21.19 & 107 \\
40 & TTM-R2 & 30 & 22.19 & 43 & 27.05 & 44 & 25.18 & 117 \\
41 & YingLong-50M & 42 & 35.16 & 37 & 24.16 & 38 & 22.50 & 117 \\
42 & YingLong-300M & 44 & 35.46 & 39 & 24.29 & 37 & 21.84 & 120 \\
43 & YingLong-110M & 43 & 35.39 & 41 & 24.47 & 40 & 23.23 & 124 \\
44 & YingLong-6M & 41 & 34.84 & 42 & 24.87 & 42 & 23.70 & 125 \\
\bottomrule
\end{NiceTabular}}
\end{table*}

\begin{figure}[t]
\centering
\adjustbox{max width=\linewidth}{\includegraphics{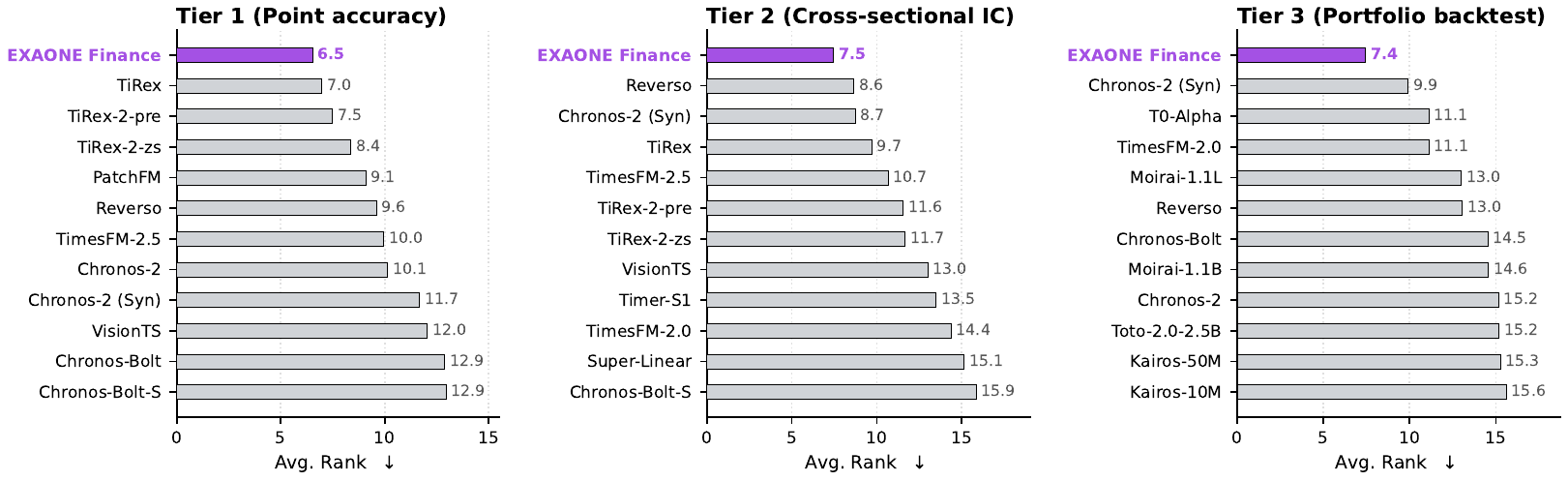}}
\caption{\textbf{Per-tier ranking.} Each panel ranks the twelve best models of one tier by
their aggregate rank (The geometric mean of per-cell ranks). \model attains the lowest
aggregate rank in all three tiers.}
\label{fig:tier}

\vspace{8pt}

\centering
\adjustbox{max width=\linewidth}{\includegraphics{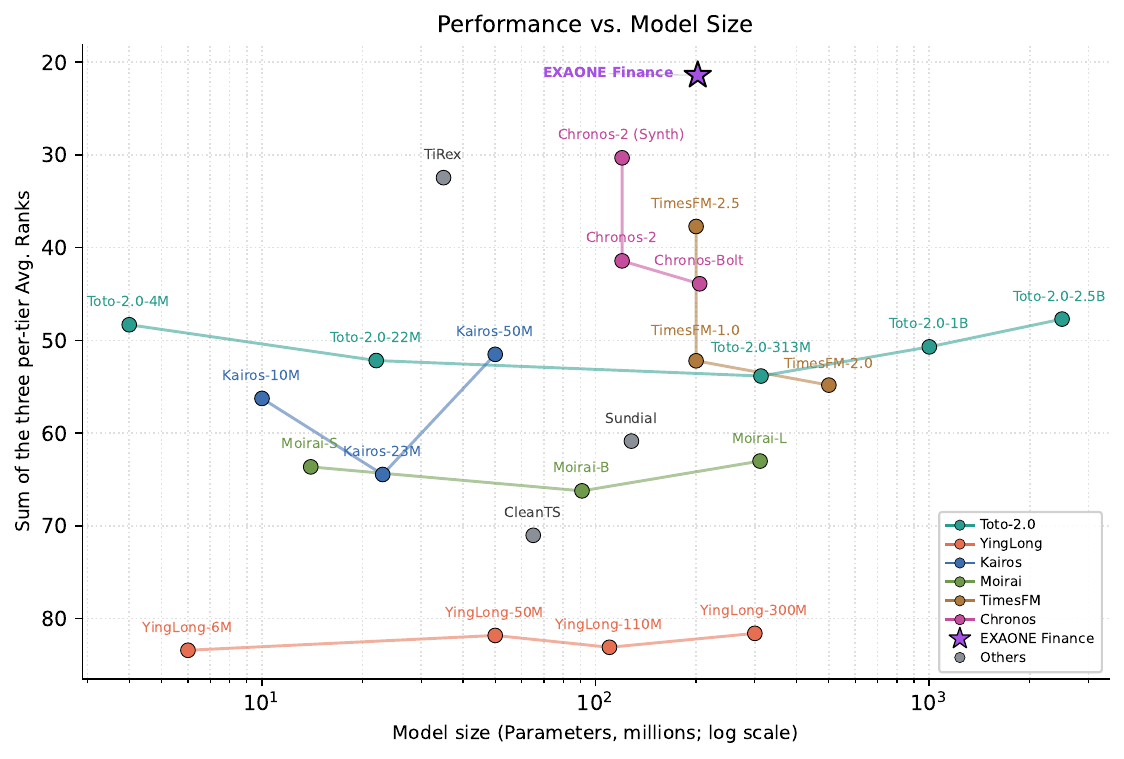}}
\caption{\textbf{Performance vs.\ model size.} Each point is a model, with parameter
count on the $x$-axis (Log scale) and, on the $y$-axis, the sum of its three per-tier
Avg.\ Ranks. Lower values are better, and the
axis is inverted so that higher is better. \model attains the best
aggregate rank at a modest size.}
\label{fig:sizeperf}
\end{figure}

\begin{figure}[t]
\centering
\adjustbox{max width=\linewidth}{\includegraphics{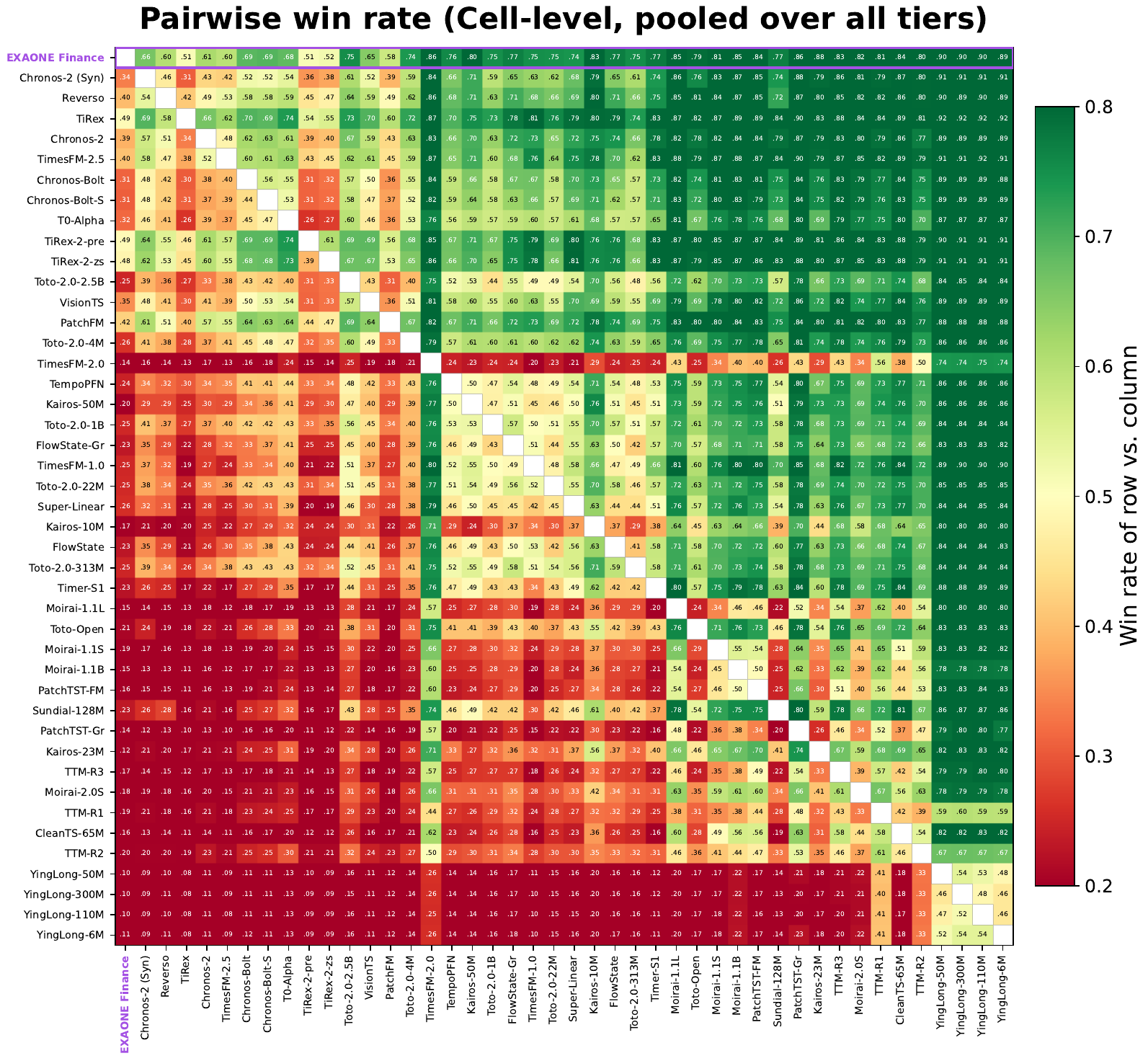}}
\caption{\textbf{Pairwise win rate.} Each cell reports the cell-level win rate of the row
model against the column model, pooled over all (tier, spec, metric) cells. Models are ordered by rank-sum, as in Table~\ref{tbl:mainB}, so \model is the first row and column. \model 
is
the only model that beats every other model head-to-head.}
\label{fig:winrate}
\end{figure}

\begin{figure*}[t]
\centering
\adjustbox{max width=\linewidth}{\includegraphics{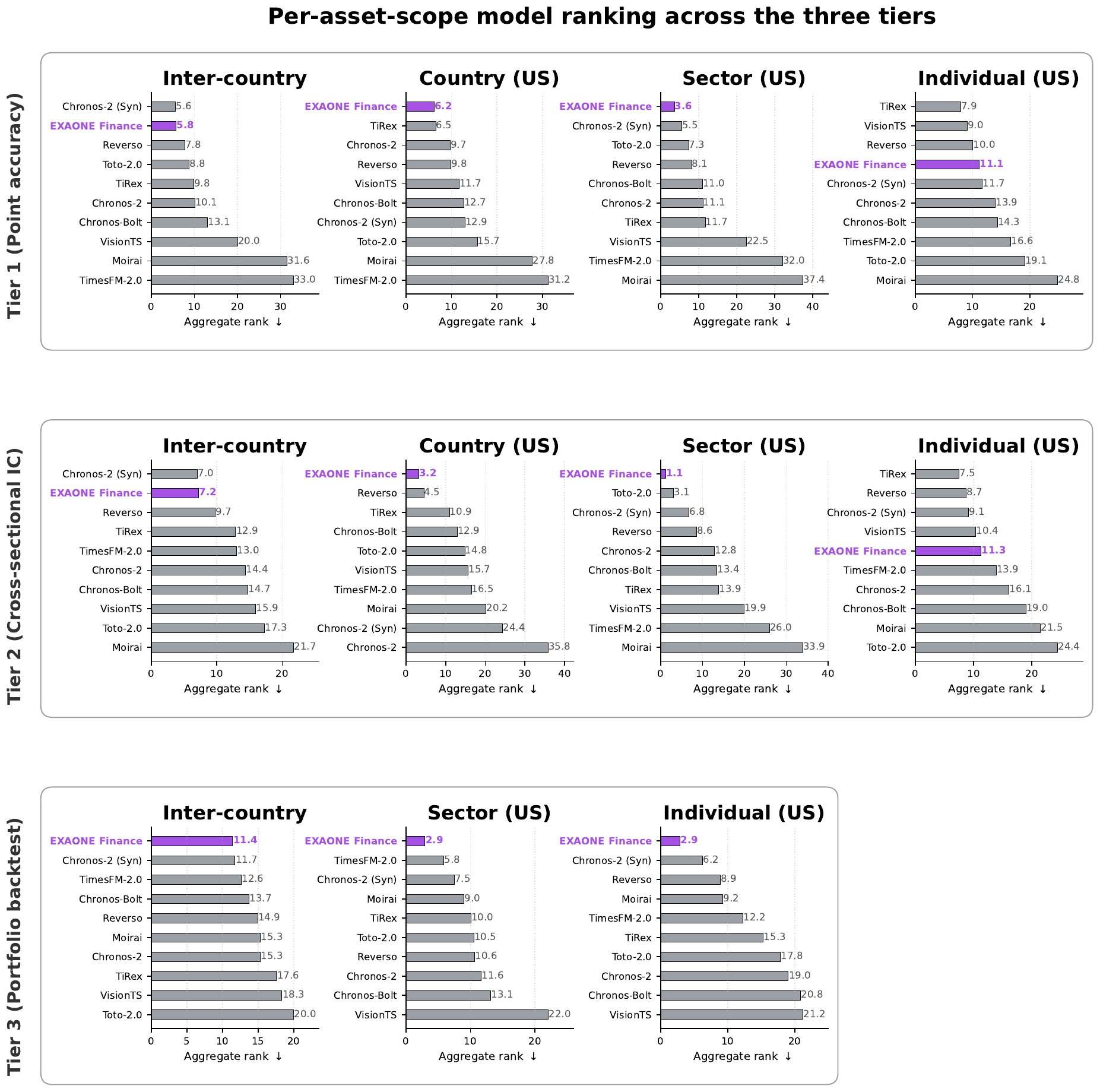}}
\caption{\textbf{Comparison of the top-10 models across the three tiers.} Each panel shows the
aggregate rank (The geometric mean of per-(Spec\,$\times$\,metric) cell ranks, where lower is
better) within the comparison set for one asset scope. The rows correspond to Tier~1 (Point
accuracy), Tier~2 (Cross-sectional IC), and Tier~3 (Portfolio backtest).}
\label{fig:bars_combined}
\end{figure*}

\subsection{Baselines and Aggregation}
\label{sec:exp_setup}
We compare against \textbf{43 baselines} covering the major public TSFM families,
including Chronos~\citep{ansari2024chronos} / Chronos-Bolt / Chronos-2~\citep{ansari2025chronos2},
Moirai (1.1, 2.0)~\citep{woo2024unified}, TimesFM (1.0/2.0/2.5)~\citep{das2024decoder},
Toto~\citep{cohen2024toto} / Toto-2.0~\citep{khwaja2026toto2}, TiRex (1.1, 2)~\citep{auer2025tirex},
TTM~\citep{ekambaram2024ttm}, PatchTST-FM~\citep{nie2023patchtst}, FlowState~\citep{graf2026flowstate},
Timer~\citep{liu2024timer}, Sundial~\citep{liu2025sundial}, Kairos~\citep{feng2025kairos},
YingLong~\citep{wang2025yinglong}, Reverso~\citep{fu2026reverso},
TempoPFN~\citep{moroshan2025tempopfn}, VisionTS~\citep{chen2025visionts}, CleanTS, PatchFM,
Super-Linear, and T0-Alpha. These span univariate
and multivariate-native APIs. Each model is scored on the same windows and ranked
with the per-cell geometric-mean-rank procedure of Section~\ref{sec:exp_protocol}
(Eq.~\ref{eq:geomean}). \model (Section~\ref{sec:model_pretrain}) is thus
evaluated against the baseline pool as a comparison set of 44 models.
\model outputs $Q=21$ quantile levels
$\{0.01,0.05,0.1,\dots,0.9,0.95,0.99\}$, and all tier metrics are computed on the
median forecast ($q_{50}$).

\subsection{Main Results}
\label{sec:exp_main}
\textbf{Overall performance.} Table~\ref{tbl:mainB} and Figure~\ref{fig:tier} present the headline result:
\emph{\model ranks \#1 in all three tiers}, for a perfect rank-sum of 3,
while the best baseline reaches only 14. The margin holds across very
different objectives, namely point accuracy (Tier 1), cross-sectional ordering (Tier 2),
and realized portfolio performance (Tier 3), indicating that the gains are not an
artifact of any single metric. Notably, no single baseline is consistently strong
across tiers. For example, TiRex~\citep{auer2025tirex} is strong on Tier 1 but drops on Tier 3, whereas \model leads every tier simultaneously. The two closest baselines, Chronos-2 (Synthetic) and Reverso (Small), never lead a tier.

\textbf{Performance vs.\ model size.}
Figure~\ref{fig:sizeperf} plots each model's aggregate rank against its parameter count,
where the aggregate is the sum of the model's three per-tier Avg.\ Ranks from
Table~\ref{tbl:mainB}.
\model sits on the Pareto frontier: at only $202$M parameters it attains the
best overall rank, outperforming models more than an order of magnitude larger
(e.g., the billion-scale Toto-2.0~\citep{khwaja2026toto2} variants), which is consistent with its
linear-time, attention-free design. The same trend appears across the other families in the figure, where a larger model size does not reliably improve the rank, suggesting that the scaling law common in general-domain forecasting does not yet hold in finance.

\textbf{Head-to-head dominance.}
To probe robustness of the ranking, we also compute the \emph{pairwise} win rate
between every pair of models: the fraction of shared (tier, spec, metric) cells on
which one model beats the other.
Ties are split evenly between the two models.
Rows and columns are ordered by rank-sum, best first, as in Table~\ref{tbl:mainB}.
As shown in Figure~\ref{fig:winrate}, \model is the
only model whose row is entirely above $0.5$, i.e., it wins its head-to-head
comparison against all 43 baselines (per-opponent win rate ranging from
$0.51$ to $0.90$).
Within that range, the narrowest margins are against the TiRex
family (TiRex-2-pretrain and TiRex-1.1, both $0.51$), the only baselines that come
close, yet both still trail \model on every tier of Table~\ref{tbl:mainB}.

\subsection{Analysis by Asset Scope}
\label{sec:exp_asset}

\textbf{Profile across all scopes.}
Because the benchmark spans a broad universe of asset classes, Figure~\ref{fig:radar}
profiles the model as a percentile over the four data
\emph{scopes} (inter-country, country, sector, individual) crossed with the three
tiers, where $100$ denotes the best of the 44 models. \model traces a
near-maximal envelope, scoring in the $95$--$100$ percentile on ten of the eleven
populated scope$\times$tier cells.

\textbf{Ranking within each scope.}
To make the comparison concrete, Figure~\ref{fig:bars_combined} breaks the ranking
down model-by-model. Each row corresponds to one tier and contains a panel per asset
scope, plotting the aggregate rank (lower is better) of \model against
representative baselines. Across point accuracy (Tier~1), cross-sectional IC
(Tier~2), and portfolio backtest (Tier~3), \model is top-3 in ten of
the eleven scope$\times$tier panels, with Chronos-2 (Synthetic), the
2nd-ranked model in Table~\ref{tbl:mainB}, as its closest competitor.

\clearpage
\begin{figure}[H]
\centering
\includegraphics[width=\linewidth]{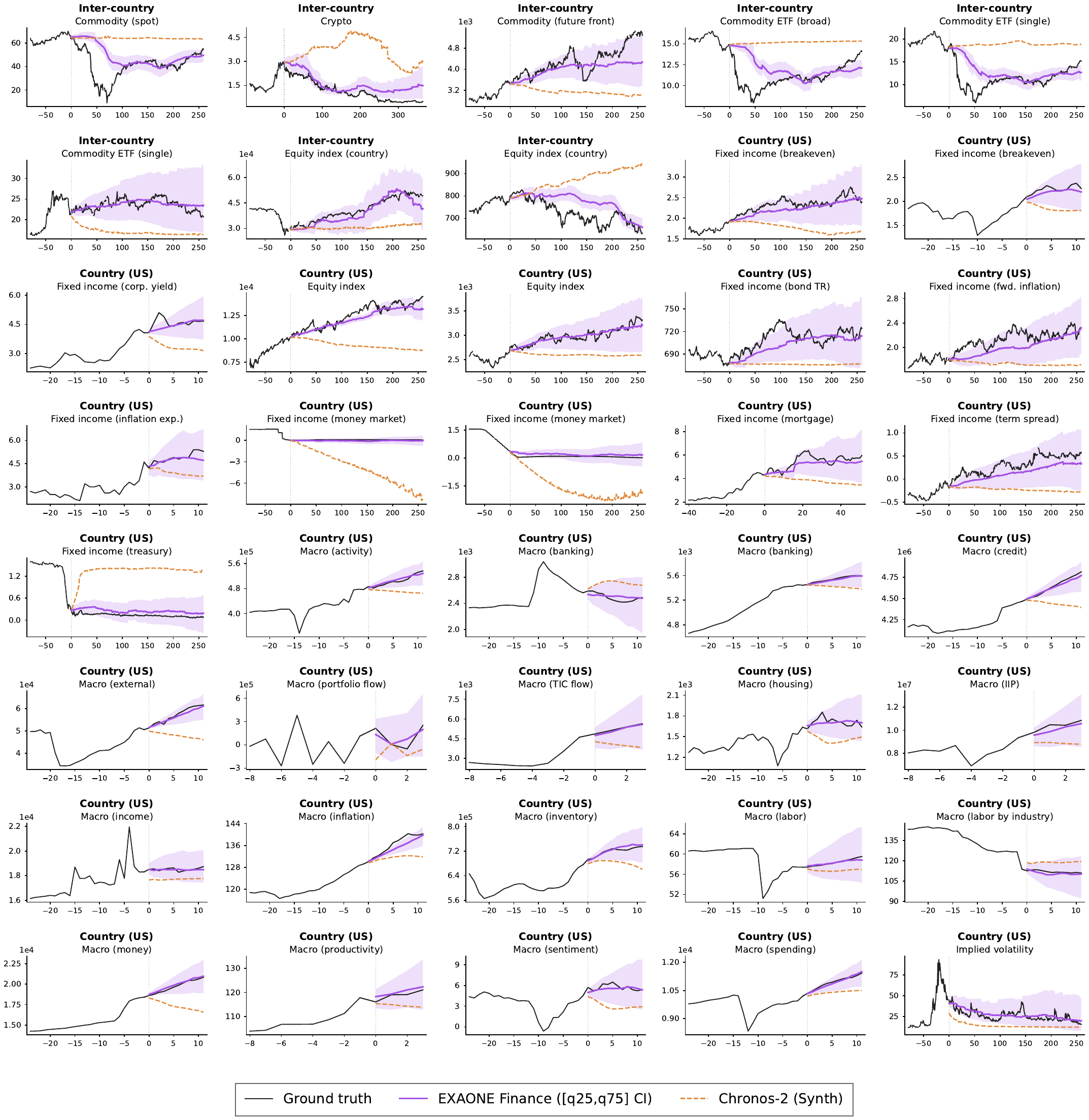}
\caption{\textbf{Forecast visualizations across asset classes (Inter-country, Country).}
Each panel shows a context history, the ground truth, the \model median
forecast with its $[q_{25},q_{75}]$ interval, and the median forecast of the 2nd-ranked
Chronos-2 (Synthetic).
\model follows the ground truth more closely than the 2nd-ranked model across
diverse sectors.}
\label{fig:examples}
\end{figure}

\begin{figure}[H]
\centering
\includegraphics[width=\linewidth]{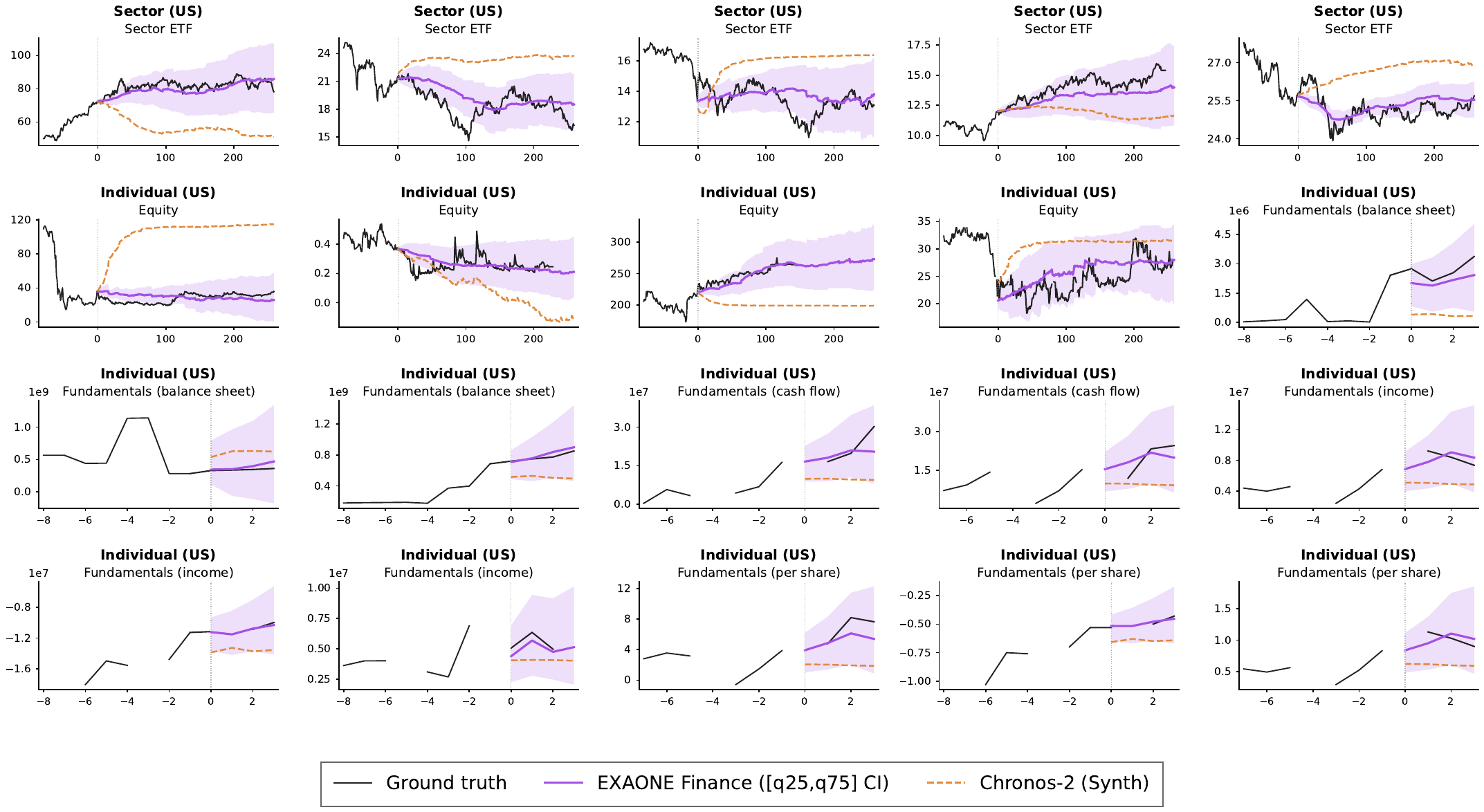}
\caption{\textbf{Forecast visualizations across asset classes (Sector, Individual).}
Same setup as Figure~\ref{fig:examples}, for the sector and individual asset scopes.}
\label{fig:examples_si}
\end{figure}

\textbf{Visualization of time series forecasts.}
Figures~\ref{fig:examples} and~\ref{fig:examples_si} visualize forecasts from \model across diverse asset classes, overlaid with the 2nd-ranked model,
Chronos-2 (Synthetic). Our median forecast (purple) tracks the ground truth more closely
than the 2nd-ranked model (orange), and the ground-truth value stays within the predicted $[q_{25},q_{75}]$
interval in most windows. Notably, several series in Figure~\ref{fig:examples_si} contain
missing spans in their context (visible as gaps in the ground-truth history), yet \model still forecasts through them accurately, consistent with its
masked-context design. Overall, these visualizations mirror the quantitative
results: \model is accurate and well calibrated across the diverse assets
of a financial panel, even under missing observations.

%% file: sections/05_conclusion.tex
\section{Conclusion}
\label{sec:conclusion}
We present \textbf{\model}, an attention-free time series foundation
model for financial forecasting. By replacing self-attention with a causal 1D
convolution for temporal mixing and a group-aware pooling MLP for variate mixing,
the model attains linear-time cost in both sequence length and variate count
while retaining a patch-based, multi-quantile decoding interface. A
masked-context augmentation makes it robust to the missingness pervasive in
financial markets. Pretrained on a \emph{synthetic financial} corpus, \model ranks \#1 in all three tiers of
the FinVerse benchmark and is the only model that wins its head-to-head comparison
against every baseline.

\textbf{Limitations and future works.}
\label{sec:limitations}
\begin{itemize}[leftmargin=1.2em,itemsep=1pt,topsep=2pt]
\item \textbf{Richer variate grouping.} The group-aware pooling mixer natively supports
arbitrary cross-series grouping structures, from the multiple fields of a single security
to related instruments within a portfolio. Exploring these grouping schemes,
and scaling to broader multivariate financial panels, is a promising direction for further
gains.
\item \textbf{Multimodal extension.} Extending the model beyond numerical series to
auxiliary modalities, such as the news, disclosures, and earnings reports that accompany a
price series, is a natural next direction~\citep{jin2024timellm,liu2024timemmd,lee2026rethinking}.
\item \textbf{Broader-domain evaluation.} Results are on a financial benchmark, and
broad general-domain benchmarking (e.g., GIFT-Eval~\citep{aksu2024gifteval}) is complementary
future work, since transfer beyond finance remains untested.
\end{itemize}